\documentclass[sigconf]{acmart}

\AtBeginDocument{%
  \providecommand\BibTeX{{%
    \normalfont B\kern-0.5em{\scshape i\kern-0.25em b}\kern-0.8em\TeX}}}

\acmYear{2021}\copyrightyear{2021}
\setcopyright{acmlicensed}
\acmConference[e-Energy '21]{The Twelfth ACM International Conference on Future Energy Systems}{June 28--July 2, 2021}{Virtual Event, Italy}
\acmBooktitle{The Twelfth ACM International Conference on Future Energy Systems (e-Energy '21), June 28--July 2, 2021, Virtual Event, Italy}
\acmPrice{15.00}
\acmDOI{10.1145/3447555.3464867}
\acmISBN{978-1-4503-8333-2/21/06}

\usepackage{lineno}
\modulolinenumbers[5]
\usepackage{url}
\usepackage{amsmath}

\usepackage[ruled,vlined]{algorithm2e}

\usepackage{multirow}
\usepackage{comment}

\usepackage{xcolor}

\begin{document}
\title{Characterizing Residential Load Patterns by Household Demographic and Socioeconomic Factors}


\author{Zhuo Wei}
\affiliation{%
  \department{Department of Data Science and AI}
  \institution{Monash University}
  \city{Melbourne}
  \state{Victoria}
  \country{Australia}
  \postcode{VIC 3800}
}
\email{zhuo.wei@monash.edu}

\author{Hao Wang}
\authornote{Corresponding author: \url{hao.wang2@monash.edu} (Hao Wang).}
\affiliation{%
  \department{Department of Data Science and AI}
  \institution{Monash University}
  \city{Melbourne}
  \state{Victoria}
  \country{Australia}
  \postcode{3800}
}
\email{hao.wang2@monash.edu}

\begin{abstract}
The wide adoption of smart meters makes residential load data available and thus improves the understanding of the energy consumption behavior. Many existing studies have focused on smart-meter data analysis, but the drivers of energy consumption behaviors are not well understood. This paper aims to characterize and estimate users' load patterns based on their demographic and socioeconomic information. We adopt the symbolic aggregate approximation (SAX) method to process the load data and use the K-Means method to extract key load patterns. We develop a deep neural network (DNN) to analyze the relationship between users' load patterns and their demographic and socioeconomic features. Using real-world load data, we validate our framework and demonstrate the connections between load patterns and household demographic and socioeconomic features. We also take two regression models as benchmarks for comparisons.
\end{abstract}




\maketitle

\section{Introduction}
Digitalization is reshaping the energy sector, and smart meters have been widely installed in households \cite{RN92}. A large amount of data collected from smart meters has great potential to enhance the energy efficiency, improve energy services, and secure system operations \cite{RN99}. Many existing studies, such as \cite{haben2015analysis}, analyzed residential customers' energy consumption behaviors using smart meter data. Studies in \cite{RN79} took a further step revealing major factors that determine energy consumption behaviors, including socioeconomic factors and dwelling factors. 
The exploration of linkages between load patterns and demographic and socioeconomic factors can help provide more personalized services. Different social groups have different load patterns \cite{jain2017data}. If these differences can be identified, there can be a better understanding of the impact of energy policies and programs on energy equity and justice.
However, the relationship between users’ energy consumption behaviors and their demographic and socioeconomic background (e.g., age, income, and education) is less well understood. Therefore, this paper aims to develop a methodology to model consumers' load patterns using smart meter data and establish the link between load patterns and demographic and socioeconomic factors.

The relationship between consumers' load patterns and their demographic and socioeconomic information has been analyzed in the literature. For example, the authors in \cite{RN3} used a random forest model to assess the impact of socioeconomic and environmental factors on residential energy consumption. The study in \cite{RN1} pointed out that the consumption behavior is more strongly linked to intrinsic factors than to characteristics of the residential environment. In \cite{RN2}, an interesting observation was made based on demographic information, and the results showed that younger users had their peak consumption later in the day than elderly users. In \cite{RN67}, socioeconomic information, such as income and education levels, was selected as features to train a prediction model for the load distribution. The studies mentioned above usually input the raw load data into clustering or classification models, and the measurement noises would affect the results. Moreover, from the operator's perspective, the information regarding load shape (e.g., single peak or dual peak) and peak time weights more than subtle load variations in the off-peak period. Hence, a more effective method is needed to model consumption behaviors using smart meter data.

This paper develops an effective methodology to address the above shortcomings in handling smart meter data and improve the understanding of the relationship between loading patterns and demographic and socioeconomic factors. 
Specifically, we use clustering techniques to extract consumers' load patterns and build a machine learning model to study the relationship between load patterns and demographic and socioeconomic factors.
\begin{itemize}
    \item We use symbolic aggregate approximation (SAX) to process the smart meter data and eliminate the effect of noises on the clustering results.
    \item We model load patterns using a set of load clusters, representing different consumption behaviors.
    \item We build a deep neural network (DNN) model to study the relationship between load patterns and demographic and socioeconomic information, such as consumers' age, income level, education level, and premise area.
    \item Our developed methodology is validated using real-world data of households.
\end{itemize}
Note that the data involved in this work are anonymized and do not involve personal privacy issues. Also, our study focuses on the energy consumption patterns at the group level, aiming to improve the understanding of different groups' behaviors for better energy equity. 

\section{Data Description}\label{sec:data}
This paper uses data from the Pecan Street database \cite{RN128} as a case study to validate our design and study the relationship between load patterns and demographic and socioeconomic factors. This database includes smart meter data of over a thousand households at a resolution of one hour. This dataset also provides these households' demographic and socioeconomic information, including the number of residents in each age group, annual household income, education level, and the household audit data. 

\subsection{Load Data}
We first pre-process the load data to remove invalid and duplicate values and obtain valid load data of $312$ households from January 1st, 2015 to December 31st, 2017. Specifically, the load data are recorded as the electricity consumption in kWh for each household in each hour, thus forming daily load profiles over three years with a total of $341,328$ data entries.

\subsection{Demographic and Socioeconomic Data}
Another essential part of the data is consumers' demographic and socioeconomic information. According to \cite{RN128}, consumers' demographic and socioeconomic information includes many aspects, such as income level, education level, age of residents, household location, and year built. Since not all the demographic and socioeconomic data serve meaningful features, simply including them will increase models' complexity and reduce accuracy. It is necessary to eliminate redundant features and select the appropriate variables. According to the results from the entropy-based feature selection in \cite{RN67}, the most relevant demographic and socioeconomic characteristics to load patterns are age, education, income, and household square footage. Therefore, we choose the above characteristics as the demographic and socioeconomic information in our study.

\section{Methods for Characterizing Load Patterns}\label{sec:method}
To study the relationship between load patterns and demographic and socioeconomic factors, we process the load data to model representative load patterns and then build a machine learning model to analyze the relationship. Our method consists of three parts: processing load data, clustering load patterns, and developing a model to find the relationship between load patterns and demographic and socioeconomic factors. These three parts are presented in Section \ref{subsec:sax}-\ref{subsec:dnn}, respectively.

\subsection{Representing Load Data with SAX}\label{subsec:sax}
We further process the original load data for the following reasons. The hourly time series load data collected by smart meters often contain measurement noises. Moreover, the operator or the utility cares more about the peak load and load type, such as dual peak or single peak and when the peaks appear. The subtle differences between load profiles, especially in off-peak hours, do not provide many insights but make load clustering less effective. So dimension reduction is necessary to eliminate these effects and capture the key information in the load profiles. In \cite{RN29}, SAX has been used to convert the load profiles into symbol strings. This process eliminates the influence of noises and subtle variations and highlights the trend change of the load profile. We use the SAX method to process the load data in our work.

The SAX algorithm proposed in \cite{RN66} extended the Piece-wise Aggregate Approximation (PAA) and kept the low complexity of the original method \cite{RN66}. It can convert time series data into different segments with a few discrete values as an approximation. In this paper, we use PAA to process the original time-series load data $\boldsymbol{L} = \{ l_1, l_2, \ldots, l_H\} $, which represents the daily load profile of $H=24$ hours. We evenly divide $24$ hours into $S$ segments, each of which has an interval length of $\tau$. PAA converts the original time-series load $\boldsymbol{L}$ into an approximating representation $\bar{\boldsymbol{L}} = \{\bar{l}_1, \bar{l}_2, \ldots, \bar{l}_S\}$, where $\bar{l}_s,~s=1,...,S$ is calculated as
\begin{equation}
\bar{l}_s=\frac{1}{\tau} \sum_{t=\tau(s-1)+1}^{\tau * S} {l_t}.
\end{equation}

After the PAA representation is complete, we use the Gaussian distribution to find interval breakpoints, each of which represents an approximation. Each segment is allocated to the approximation corresponding to its closest break-point, thus dividing the time-series load profile into a few segments with several discrete values \cite{RN66}. In our case study, we choose $\tau = 3$ and divide the $24$-hour load profiles into $S = 8$ segments. The load data in each segment can be represented by five discrete values \cite{RN125} as an approximation of consumers' load profiles.

\subsection{Modeling Load Patterns via Clustering}\label{subsec:cluster}
After processing the load data with SAX, it is necessary to extract several representative load patterns from the processed load data. Specifically, we use the clustering technique to model load patterns by dividing the SAX-processed load profiles into several clusters, each of which represents a representative load pattern.

The partitioning clustering methods \cite{RN97} have been widely used to cluster time-series data, and K-means is one of the most commonly used partitioning clustering methods. The main advantage of K-means is that it is easy to implement and scale-up \cite{RN114}, and thus we choose K-means as the clustering method in our work. The number of clusters $K$ must be chosen carefully to represent typical load patterns. If $K$ is too large, clusters can be very similar to each other and do not exhibit meaningful differences. If $K$ is too small, some typical load patterns may be missing. Therefore, we consider both Sum of Squared Errors (SSE) and the silhouette index \cite{RN127} to select the most appropriate value for $K$.

A smaller SSE indicates better clustering performance. As the number of clusters $K$ increases, the SSE will gradually become smaller. When $K$ reaches an appropriate number,  the declining rate of SSE diminishes and stabilizes even with an increasing $K$. In our case, the SSE starts to stabilize after the cluster number $K$ reaches $5$, so any $K$ greater than $5$ can be considered. To determine the best $K$, we further consider the silhouette index, as it is also an important measure of clustering performance \cite{RN127}. A larger silhouette index means that the load profiles are closer to each other within a cluster but more distant from each other across different clusters, indicating better clustering performance. In our study, the silhouette index peaks at a cluster number of $K = 7$. Therefore, we choose $K= 7$ clusters to model load patterns for achieving a low SSE and the highest silhouette index.

To sum up, this paper divides the load profiles into $7$ clusters by the K-means method, representing $7$ different electricity consumption behaviors. We further study how the consumers' load patterns are correlated with the demographic and socioeconomic features in Section~\ref{subsec:dnn}.

\subsection{Machine Learning Model Based on DNN}\label{subsec:dnn}
We develop a machine learning model to explore the relationship between demographic and socioeconomic characteristics and consumers' load patterns. Our problem can be regarded as a regression problem, in which the independent variables are consumers' demographic and socioeconomic features, and the dependent variables are the probabilities of load patterns. We tried to solve the problem with regression models, such as linear regression and polynomial regression, but the fit was not satisfactory due to unknown nonlinearity in the data. Therefore, we tried more efficient models. DNN is a powerful machine learning method with many successful applications \cite{RN67}. It consists of a collection of neurons organized in a sequence of multiple layers, which can incorporate nonlinear activation functions. Therefore, this paper uses a DNN model and takes a linear regression model and a polynomial regression model as two benchmark models.

After the clusters have been identified in our study, we have a set of representative load patterns. Each consumer's load profiles can be modeled as a probability distribution over the representative load patterns. We take users' demographic and socioeconomic features as the input of the DNN model to find the relationship between them and the load patterns. 
A set of percentages of load profiles belonging to each representative load pattern is the estimated target, i.e., the DNN model's output. 
The probability for each cluster is always in the range [0,1], and the sum of all probabilities is always $1$. Therefore, we introduce the SoftMax function to normalize the output as a valid probability distribution.

In summary, the developed DNN based model takes consumers' demographic and socioeconomic features as the input and the probability distribution of load patterns as the output, thereby correlating the demographic and socioeconomic features with load patterns and revealing their relationship.

\section{Case Study}
To validate our developed method in Section~\ref{sec:method}, we use real-world data described in Section~\ref{sec:data} as a case study. We implement a DNN model with $3$ hidden layers, each of which includes $100$ neurons. The number of residents in each age group, annual household income, the education level of the residents, and total square footage of the house are used as inputs. The outputs are the probability distributions of load patterns. Considering the foreseeable difference in behaviors on weekdays and weekends, we split the load data into weekday and weekend load data. We use 80\% of the data for training and the remaining 20\% for the test based on the weekday and weekend load data, respectively. 

\subsection{Model Performance}
\begin{figure*}[!htbp] %
    \centering
    \includegraphics[width=0.85\linewidth]{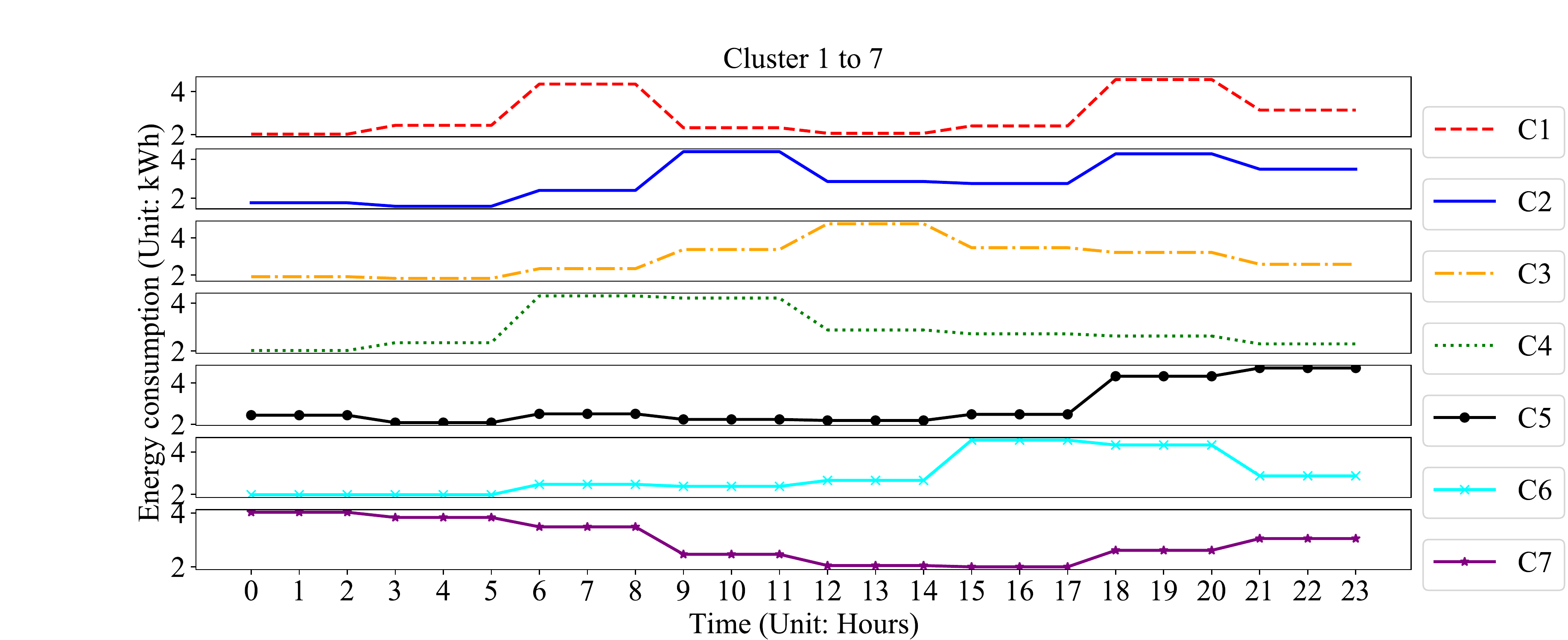}
    \caption{The $7$ representative load patterns on weekdays.}
    \label{f12}
\end{figure*}

\begin{figure*}[!htbp] %
    \centering
    \includegraphics[width=0.32\linewidth]{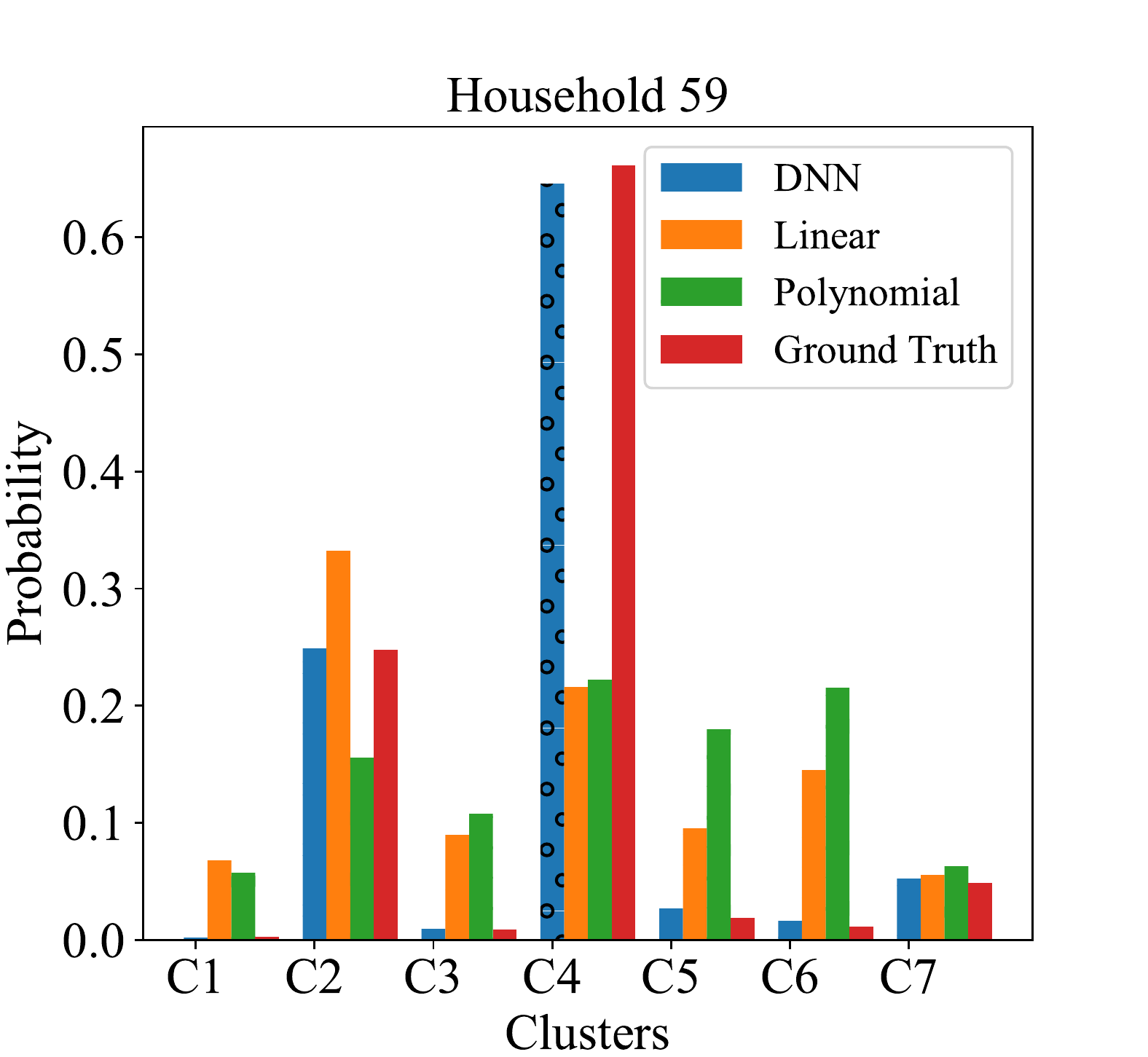}
    \includegraphics[width=0.32\linewidth]{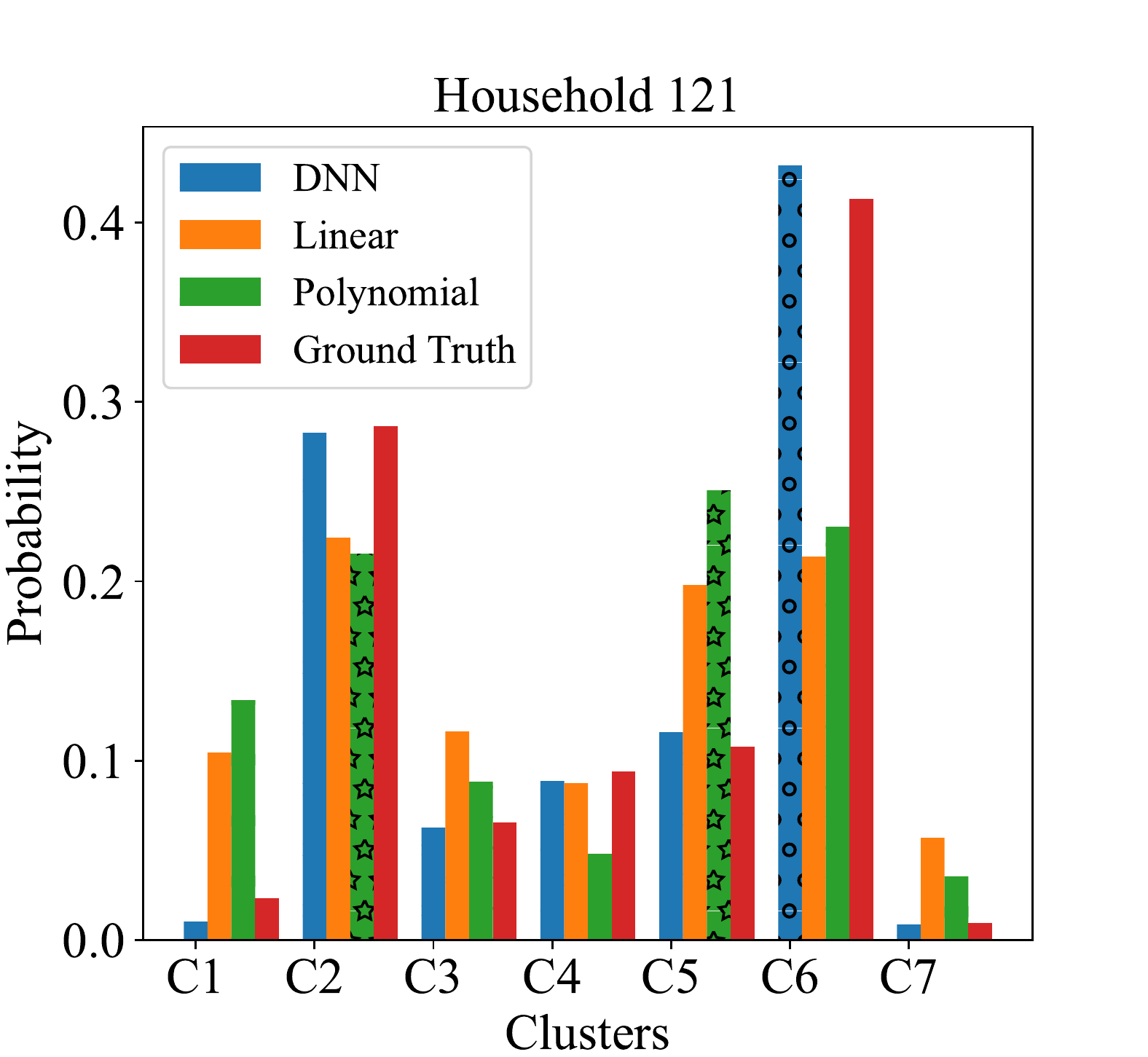}
    \includegraphics[width=0.32\linewidth]{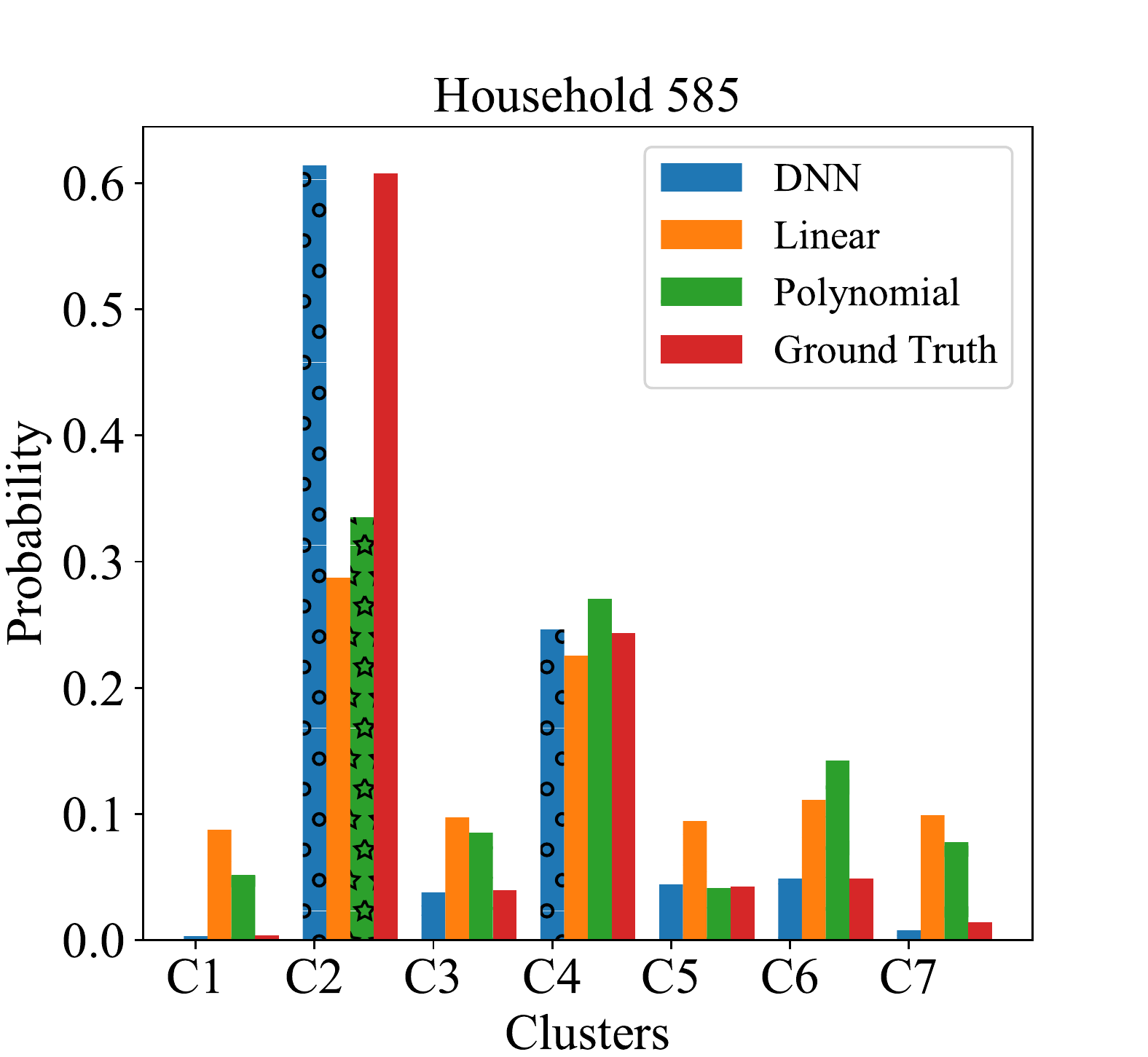}
    \caption{The comparison of three different users on weekdays.}
    \label{f11}
\end{figure*}

We compare our developed DNN model with a linear regression model and a polynomial regression model. The data used for training and test are the same for all of the models. Due to the page limit, we only depict the comparison results for weekday load. Figure \ref{f12} shows the load patterns described as $7$ clusters. Figure \ref{f11} shows a comparison of weekday load patterns for three typical consumers with different demographic and socioeconomic features.

In Figure \ref{f12}, we see that the lower electricity consumption periods may suggest that residents are resting or not at home, while the first peak load (as shown in Clusters 1, 2, and 4) usually means the start of activities in the morning, such as preparing breakfast or doing laundry before going to work. In some clusters (such as Clusters 1, 2, 5, and 6), we see peak load in the evening, suggesting that residents get back home after work or start preparing dinner and have recreational activities. In Cluster 3, the peak occurs during midday, which indicates that residents are at home and residents' activities are concentrated throughout the day. Cluster 7 has a late-night peak, but the electricity consumption is very low during the day. A possible reason for this load pattern is that residents are sleeping or not at home during the daytime but awake at night.

The residents' load patterns reflect their household activities, which are correlated with their demographic and socioeconomic background. Figure \ref{f11} shows the estimated results on three residents' probability distributions of load patterns based on their demographic and socioeconomic features using our DNN model and two regression benchmarks, compared with the ground truth. We see that the probability distributions of the load patterns for both Household 59 and Household 585 are concentrated on two specific clusters, but the probability distribution of the load patterns for household 121 is less concentrated, which may be related to the age distribution of the residents. Both Household 59 and Household 585 have residents with ages over 65, but Household 121 also has children under 12 years of age, which reveals that the elderly usually have a more regular routine. Moreover, both Household 59 and Household 585 have load patterns concentrated in Cluster 2 and Cluster 4 with a peak during morning hours. This could be because elderly users are more inclined to wake up early and have a hot breakfast. The probability distribution of Household 121 is less concentrated than those of the other two households. But Household 121 has a higher fraction of load in Cluster 2 and Cluster 6, indicating a peak during the evening hours. This can be caused by households with children spending the evening for some family entertainment activities. These households having peak load in the evening are more likely to be called for demand response, and their household activities will be affected.

Furthermore, the DNN model has a better fit than the two benchmark models. Due to the complex relationship between consumers' demographic and socioeconomic characteristics and the probability distribution of their load patterns, the two benchmark models may under-fit, while our DNN model can provide more accurate results. The errors of the three models are measured by the Mean Squared Error (MSE), and a smaller MSE means better fitting. The DNN model reduces MSE by 90\% compared to the linear regression model and more than 70\% compared to the polynomial regression model. Note that different results will be obtained using a different dataset. Our work provides a methodology to model the load patterns and study the relationship between load patterns and demographic and socioeconomic features. We have no intention of emphasizing the improvement in the estimation accuracy.

\section{Discussions and Conclusion}
We developed an analytical method to model the residential load patterns and discover the relationship between the load patterns and the demographic and socioeconomic factors. We use the SAX method to remove the impact of measurement noises in load data, symbolically represent the data and approximate the load profiles. We use the K-means clustering to model the representative load patterns. Using demographic and socioeconomic factors as inputs and load patterns as outputs, we develop a DNN model to study the relationship between the load data and demographic and socioeconomic data. Our model generates new insights into how consumers with different demographic backgrounds use electricity.

In our future work, we plan to reduce the impact of measurement noises in the load data to model load patterns. The data availability also restricts the current results, and we plan to find more a large dataset for validation. We also aim to analyze the impact of energy tariffs and policy on consumers and improve energy equity.

\newpage
\bibliographystyle{ACM-Reference-Format}
\bibliography{ref.bib}

\end{document}